\documentclass[runningheads]{llncs}
\usepackage{graphicx}
\usepackage{amsmath}
\usepackage{url}
\usepackage{hyperref}
\usepackage{xcolor}
\usepackage{graphicx}
\usepackage[caption=false]{subfig}
\usepackage{array}
\usepackage{xcolor}

\begin{document}
\title{A Continuous Optimisation Benchmark Suite from Neural Network Regression}
%
%
\author{Katherine M. Malan \inst{1}\orcidID{0000-0002-6070-2632} \and
Christopher W. Cleghorn\inst{2}\orcidID{0000-0002-7860-0650}}

\authorrunning{Malan and Cleghorn}
%
\institute{Department of Decision Sciences, University of South Africa, Pretoria, South Africa\\ \email{malankm@unisa.ac.za} \and
School of Computer Science and Applied Mathematics, University of the Witwatersrand, Johannesburg, South Africa\\
\email{christopher.cleghorn@wits.ac.za}}
\maketitle              
\begin{abstract}
Designing optimisation algorithms that perform well in general requires experimentation on a range of diverse problems. Training neural networks is an optimisation task that has gained prominence with the recent successes of deep learning. Although evolutionary algorithms have been used for training neural networks, gradient descent variants are by far the most common choice with their trusted good performance on large-scale machine learning tasks. With this paper we contribute CORNN (Continuous Optimisation of Regression tasks using Neural Networks), a large suite for benchmarking the performance of any continuous black-box algorithm on neural network training problems. Using a range of regression problems and neural network architectures, problem instances with different dimensions and levels of difficulty can be created. We demonstrate the use of the CORNN Suite by comparing the performance of three evolutionary and swarm-based algorithms on over 300 problem instances, showing evidence of performance complementarity between the algorithms. As a baseline, the performance of the best population-based algorithm is benchmarked against a gradient-based approach. The CORNN suite is shared as a public web repository to facilitate easy integration with existing benchmarking platforms.

\end{abstract}
\section{Introduction}
The importance of a good set of benchmark problem instances is a critical component of a meaningful benchmarking study in optimisation~\cite{BART2020}. As a consequence of the No Free Lunch Theorems for optimisation~\cite{WOLP1997}, if an algorithm is tuned to improve its performance on one class of problems it will most likely perform worse on other problems~\cite{HAFT2016}. 
Therefore, to develop algorithms that perform well in general or are able to adapt to new scenarios, a wide range of different problem instances are needed for experimental algorithm design.   

Most of the benchmark problems available for continuous optimisation are artificial, so performance achieved through tuning algorithms on these problems cannot be assumed to transfer to real-world problems. On the other hand, testing algorithms on real-world problems is not always feasible and has the disadvantage of not covering the wide range of problem characteristics needed for different problem scenarios. To bridge this gap, we introduce a benchmark suite from the real-world domain of neural network (NN) training that includes some of the advantages of artificial benchmark problems. 

Metaheuristics frequently suffer from the {\it curse of dimensionality} with performance degrading as the number of decision variables increases~\cite{LOZA2010,MAHD2015,OLDE2017,TANG2009}. This is in part due to the increased complexity of the problem, but also due to the exponential growth in the size of the search space~\cite{LOZA2010}. The training of NNs presents an ideal context for high-dimensional optimisation as even a medium-sized network will have hundreds of weights to optimise. Most NN training studies use a limited number of problem instances (classification datasets or regression problems), which brings into question the generalisability of the results. For example, in six studies using population-based algorithms for NN training~\cite{DAS_2014,MIRJ2012,MIRJ2015,MOUS2020,SOCH2007,VAND2000}, the number of problem instances used for testing ranged from a single real-world instance~\cite{DAS_2014} to eight classification or regression problems~\cite{MIRJ2015,MOUS2020}. To facilitate the generalisability of NN training studies, we provide a suite of hundreds of problem instances that can easily be re-used for benchmarking algorithm performance. 

Stochastic gradient descent~\cite{LECUN2015} is the default approach to training NNs with its trusted good performance on large-scale learning problems~\cite{BOTT2012}. Population-based algorithms have been proposed for training NNs~\cite{DAS_2014,MIRJ2012,MIRJ2015,MOUS2020,SOCH2007,VAND2000}, but they are seldom used in practice. One of the challenges is that the search space of NN weights is unbounded and algorithms such as particle swarm optimisation may fail due to high weight magnitudes leading to hidden unit saturation~\cite{RAKI2014,RAKI2015}. 
The one domain where population-based metaheuristics have shown competitive results compared to gradient-based methods is in deep reinforcement learning tasks~\cite{SUCH2018}. More benchmarking of gradient-free methods against gradient-based methods is needed to highlight the possible benefits of different approaches.

This paper proposes CORNN, Continuous Optimisation of Regression tasks using Neural Networks, a software repository of problem instances, including benchmarking of population-based algorithms against a gradient-based algorithm (Adam). Over 300 regression tasks are formed from 54 regression functions with different network architectures generating problem instances with dimensions ranging from 41 to 481.  The source code and datasets associated with the benchmark suite are publicly available at \href{https://github.com/CWCleghornAI/CORNN}{github.com/CWCleghornAI/CORNN}.

\section{Continuous Optimisation Benchmark Suites}
When benchmarking algorithms, it is usually not practical to use real-world problems, due to the limited range of problem instances available and the domain knowledge required in constructing the problems. Individual real-world problem instances also do not effectively test the limits of an algorithm because they will not usually cover all the problem characteristics of interest~\cite{RARD2001}. Artificial benchmark suites and problem instance generators have therefore become popular alternatives for testing optimisation algorithms. 
In the continuous optimisation domain, the most commonly used artificial benchmark suites include the ACM Genetic and Evolutionary Computation Conference (GECCO) BBOB suites~\cite{BBOB} and IEEE Congress on Evolutionary Computation (CEC) suites~\cite{CEC}. 
These artificial suites have been criticised for having no direct link to real-world settings~\cite{FISC2020}, resulting in a disconnect between the performance of algorithms on benchmarks and real-world problems~\cite{TANG2019}.

To address the limitations of artificial benchmarks, suites that are based on real-world problems or involve tasks that are closer to real-world problems have been proposed, such as from the domains of electroencephalography (EEG)~\cite{Goh2015}, clustering~\cite{GALL2016}, 
crane boom design~\cite{FLEC2019}, and games~\cite{VOLZ2019}. The CORNN Suite proposed in this paper extends these sets to include the class of problems for solving NN regression tasks. These tasks are unique in that the decision variables are unbounded and the scenario makes it possible to benchmark black-box algorithms against gradient-based approaches.

\section{Neural Network Training Landscapes}
Training NNs involves adjusting weights on the connections between neurons to minimise the error of the network on some machine learning task. Since weight values are real numbers, the search space is continuous with dimension equal to the number of adjustable weights in the network. Training NNs is known to be NP-complete even for very small networks~\cite{BLUM1992} and the properties of error landscapes are still poorly understood~\cite{CHOR2015} with conflicting theoretical claims on the presence and number of local minima~\cite{AUER1996,BALD1989,HAME1998,MEHT2018}. Some studies have suggested that these landscapes have large flat areas with valleys that radiate outwards~\cite{CHAU2019,GALL2000,KESK2017,KORD2004} and a prevalence of saddle points rather than local minima~\cite{CHOR2015,DAUP2014}. Saddle points present a challenge for search, because they are generally surrounded by high error plateaus~\cite{DAUP2014} and, being stationary points, can create the illusion of being local optima. It has, however, also been found that failure of gradient-based deep learning is not necessarily related to an abundance of saddle points, but rather to aspects such as the level of informativeness of gradients, signal-to-noise ratios and flatness in activation functions~\cite{SHAL2017}.

To better understand the nature of NN error landscapes, investigations are needed into the behaviour of different algorithms on a wide range of problems. The CORNN Suite can be used to complement existing suites or as a starting point for this kind of analysis. 
\section{The CORNN Benchmark Suite}

With sufficient neurons, NNs are able to model an arbitrary mathematical function~\cite{BISH1995,HUAN2006}, so are a  suitable model for solving complex regression problems. The optimisation task of the CORNN Suite involves fitting a fully-connected feed-forward NN to a real-valued function, $f(\mathbf{x})$. Each network has an $n$-dimensional real-valued input,$\mathbf{x}$, and a single real-valued output, which is the prediction of the target value, $f(\mathbf{x})$. The CORNN Suite uses 54 two-dimensional functions as the basis for regression fitting tasks. These functions are specified on the CORNN Suite repository on github \footnote{\url{https://github.com/CWCleghornAI/CORNN/blob/main/CORNN_RegressionFunctions.pdf}}. The functions cover a range of characteristics with respect to modality, separability, differentiability, ruggedness, and so on. Note, however, that although these characteristics will no doubt have some effect on the difficulty of the regression task, we cannot assume that the features of the functions relate to the characteristics of the higher dimensional search space of NN weights for fitting the functions.

\subsection{Training and Test Sets}
Datasets were generated for each of the 54 functions as follows: 5000 $(x_1,x_2)$ pairs were sampled from a uniform random distribution of the function's domain. A value of 5000 was used to be large enough to represent the actual function, while still allowing for reasonable computational time for simulation runs. The true output for each $(x_1, x_2)$ pair was calculated using the mathematical function and stored as the target variable. Each dataset was split randomly into training (75\% of samples) and testing (25\% of samples) sets. Two forms of preprocessing were performed on the CORNN Suite datasets:(1) Input values were normalised to the range $[-1, 1]$ using the domain of each function; (2) To compare results of problem instances with different output ranges, output values were normalised using simple min-max scaling based on the training data to the range $(0, 1)$. 

In addition, the CORNN Suite's implementation allows for the use of custom datasets; either generated from analytic functions or existing datasets. 

\subsection{Neural Network Models}
The architecture used in the CORNN Suite is a fully connected feed-forward network with 2 inputs; 1-, 3-, or 5-hidden layers, each with 10 neurons plus a bias unit; and 1 output neuron. This results in 41, 261, and 481 weights to be optimised for the 1-, 3-, and 5-layer networks respectively. Each architecture uses one of two hidden layer activation functions: the conventional hyperbolic tangent (Tanh) and the rectified linear unit (ReLU). ReLU is currently the most commonly used activation function in deep learning~\cite{LECUN2015}, but Tanh has been recommended above ReLU for reinforcement learning tasks~\cite{ANDR2020}. The output layer uses a linear activation function in all cases. These six topologies are referred to as Tanh1, Tanh3, Tanh5, ReLU1, ReLU3, and ReLU5, specifying the activation function and number of hidden layers. The CORNN Suite therefore consists of $54 \times 6 = 324$  problem instances, since each function has six NN models for fitting the function. Note that the CORNN Suite's implementation allows for complete customisation of architectures to create any desired topology for further analysis. 

\subsection{Performance Evaluation}
Performance of an algorithm is measured using mean squared error (MSE) of the trained model on the test set given a set budget of function evaluations. Note that in the analysis presented in this paper, no evidence of overfitting was observed. If overfitting becomes a consideration as more specialised optimisers are developed/considered, it may become necessary to hold out a portion of the training set to employ techniques such as early stopping etc. When using hold-out training instances, the total number of function evaluations should be seen as the maximum number of times any one training instance has been used. If no hold-out instances are used this measurement is equivalent to the number of full passes of the training data. A similar consideration should be made if an optimiser requires hyper-parameter tuning; in such cases hold-out instances from the training set should be used for tuning and not the test set instances.

\subsection{Implementation Details}
The CORNN Suite was developed in Python 3 using PyTorch.
The user selects a regression task and a model architecture, after which the library constructs a problem instance object with a callable function to which the user can pass a candidate solution for evaluation on either the training set during optimisation, or on the test set after optimisation. A user of CORNN therefore does not have to concern themselves with any data processing or NN computation. The complexity of the problem instances are abstracted away to the point where a user of CORNN can just work with an objective function after setup.

The GitHub repository provides installation instructions with a detailed example of how to construct and use a CORNN problem instance. The suite is easily extended beyond the $324$ problem instances presented in this paper to include other regression tasks and/or NN architectures through reflection. The datasets for all $54$ regression tasks are also provided in CSV format, but when using the CORNN Suite it is not necessary to directly interact with these files.

\section{Experimentation and Results}
To demonstrate CORNN, we provide results of metaheuristics and contrast these with a gradient-based method. The aim is not to compare algorithms, but to provide a use-case of the suite. No tuning of algorithm parameters was done, so the results are not representative of the best performance of the algorithms. 

\subsection{Experimental Setup}
The algorithms used in this study were: particle swarm optimisation (PSO)~\cite{KENN1995}, differential evolution (DE)~\cite{STORN1997}, covariance matrix adaptation evolution strategy (CMA-ES)~\cite{HANS1996}, Adam~\cite{ADAM2014}, and random search. For the population-based algorithms, standard versions and hyperparameters defined in the Nevergrad~\cite{BENN2021} library were used to facilitate reproducibility\footnote{Nevergrad version 0.4.2, PSO with \textit{optimizers.PSO}, DE with \textit{optimizers.TwoPointsDE}, CMA-ES with \textit{optimizers.CMA}, and random search with \textit{optimizers.RandomSearch}}. Adam was used for the gradient-based approach (PyTorch \cite{PyTorch2019} implementation with default parameters). Each algorithm had a function evaluation (FE) budget of $5000$ per problem instance, where an FE is defined as one complete pass through the training dataset. We used full batch learning, but the suite is not limited to this approach. Optimisation runs were repeated $30$ times for each algorithm/problem instance pair. 

\subsection{Analysis of Population-based Algorithms}
\label{sec:algperf}
The first set of results contrasts the performance of the three population-based algorithms: CMA-ES, DE and PSO against random search. We only present the performance on the testing datasets, because we found no evidence of overfitting by any of the algorithms on the problem set. Each algorithm is given a performance score at each evaluation using the following scoring mechanism per problem instance against each competing algorithm:
\begin{itemize}
    \item 1 point is awarded for a draw (when there is no statistically significant difference based on a two-tailed Mann-Whitney U test with 95\% confidence).
    \item 3 points are awarded for a win and 0 for a loss. In the absence of a draw, we determine whether a win or loss occurred using one-tailed Mann-Whitney U tests (with 95\% confidence). 
\end{itemize}
This results in a maximum score for an algorithm on a single instance of 9, or $3(n - 1)$ in general, where $n$ is the number of  algorithms. The scores per instance were normalised to the range $[0,1]$.

\begin{figure}[!t]
\begin{center}
	\subfloat[1-layer Tanh model]{%
		\includegraphics[width=0.49\textwidth]{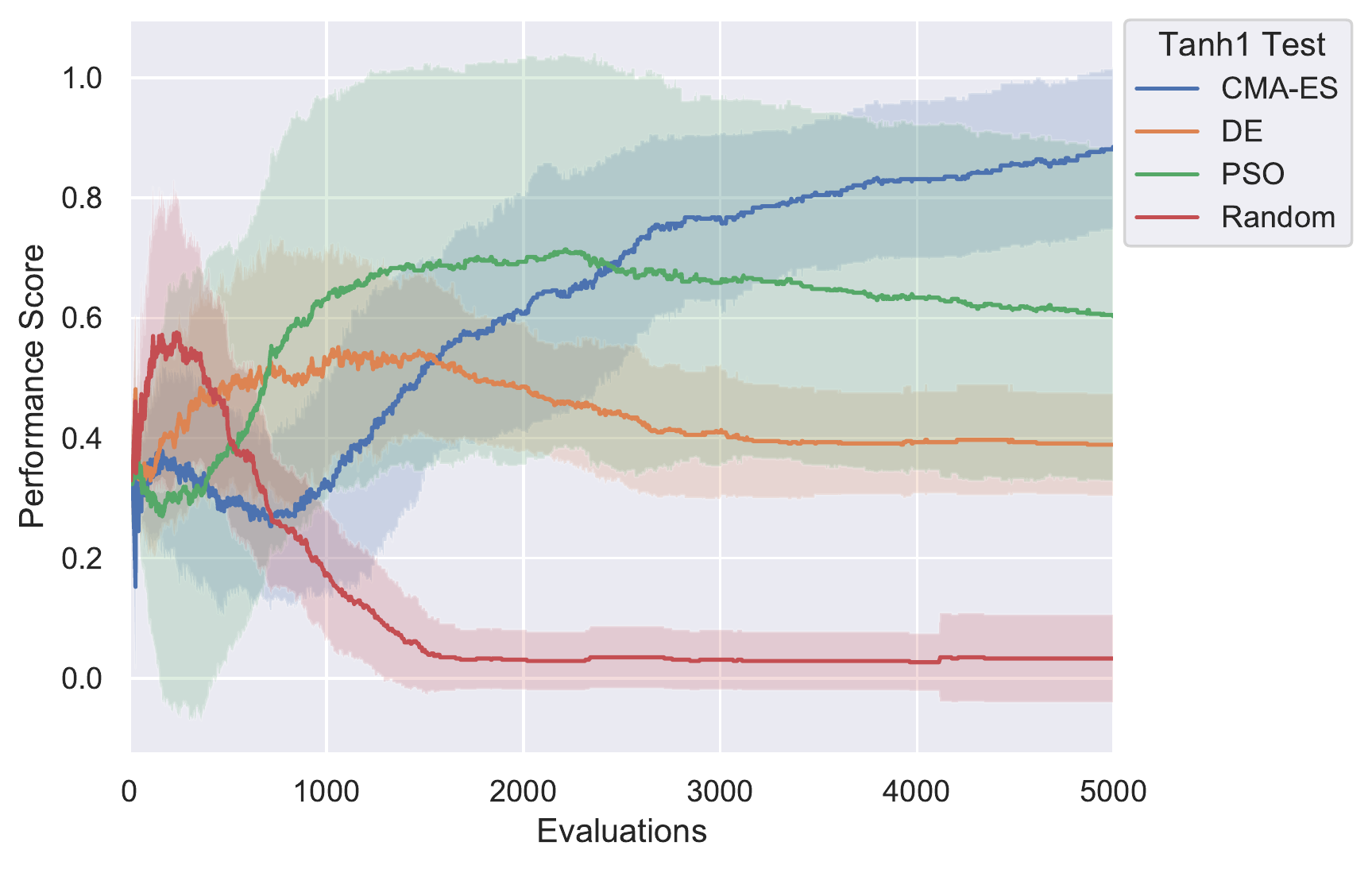}
	}
	\hfill
		\subfloat[1-layer ReLU model]{%
		\includegraphics[width=0.49\textwidth]{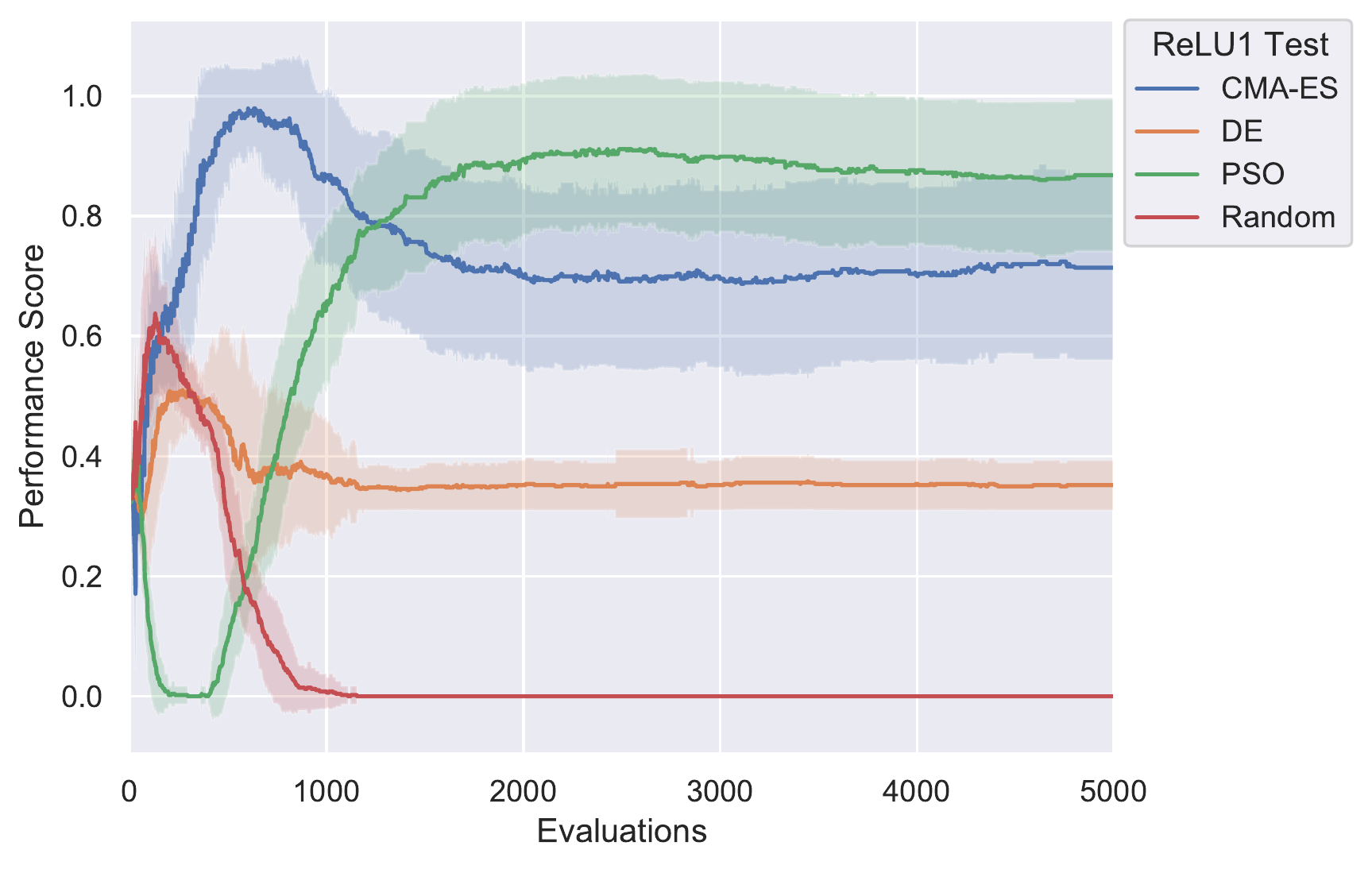}
	}\\
	\subfloat[3-layer Tanh model]{%
		\includegraphics[width=0.49\textwidth]{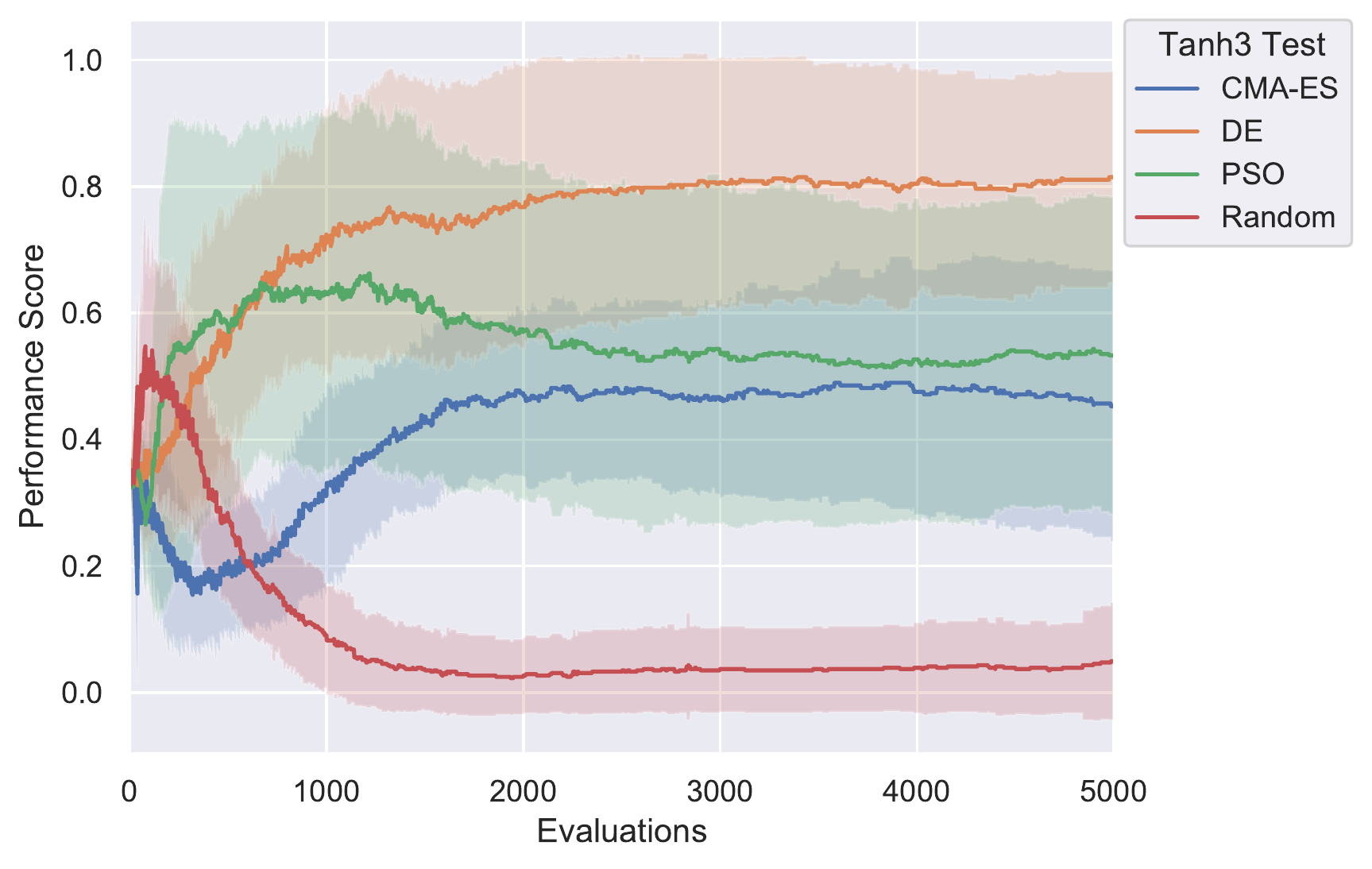}
	}
		\subfloat[3-layer ReLU model]{%
		\includegraphics[width=0.49\textwidth]{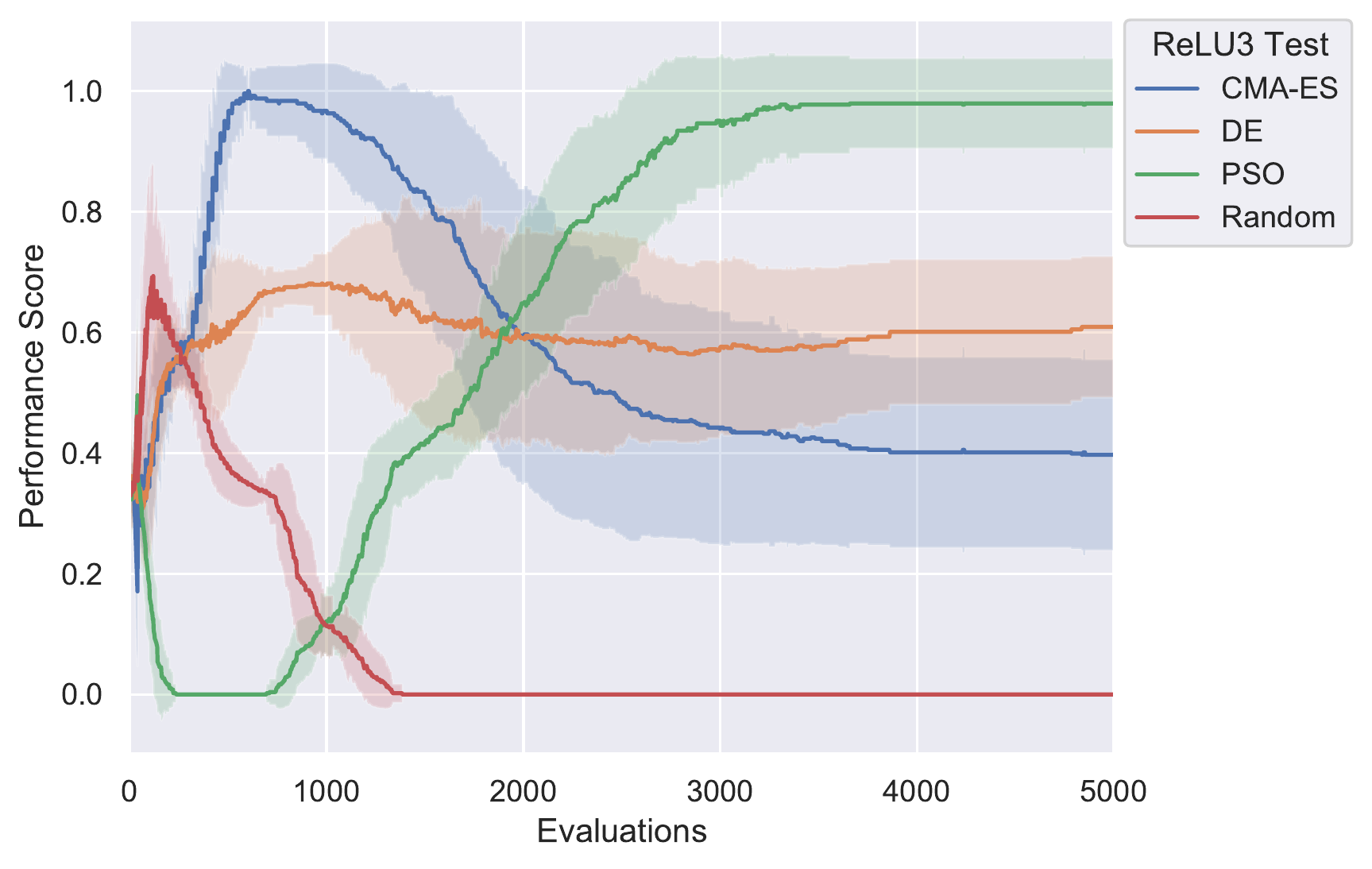}
	}\\	
	\subfloat[5-layer Tanh model]{%
		\includegraphics[width=0.49\textwidth]{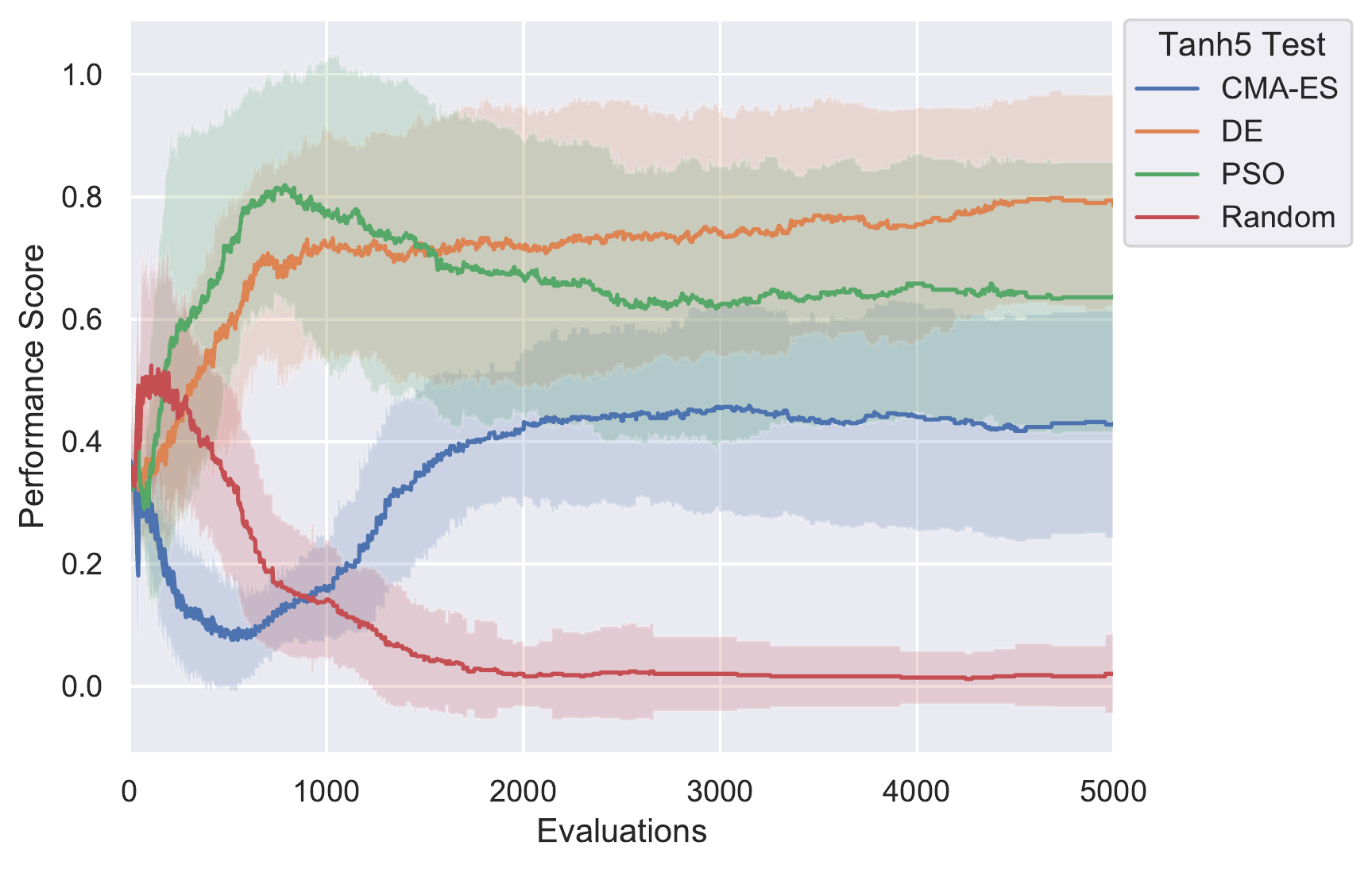}
	}
	\subfloat[5-layer ReLU model]{%
	\includegraphics[width=0.49\textwidth]{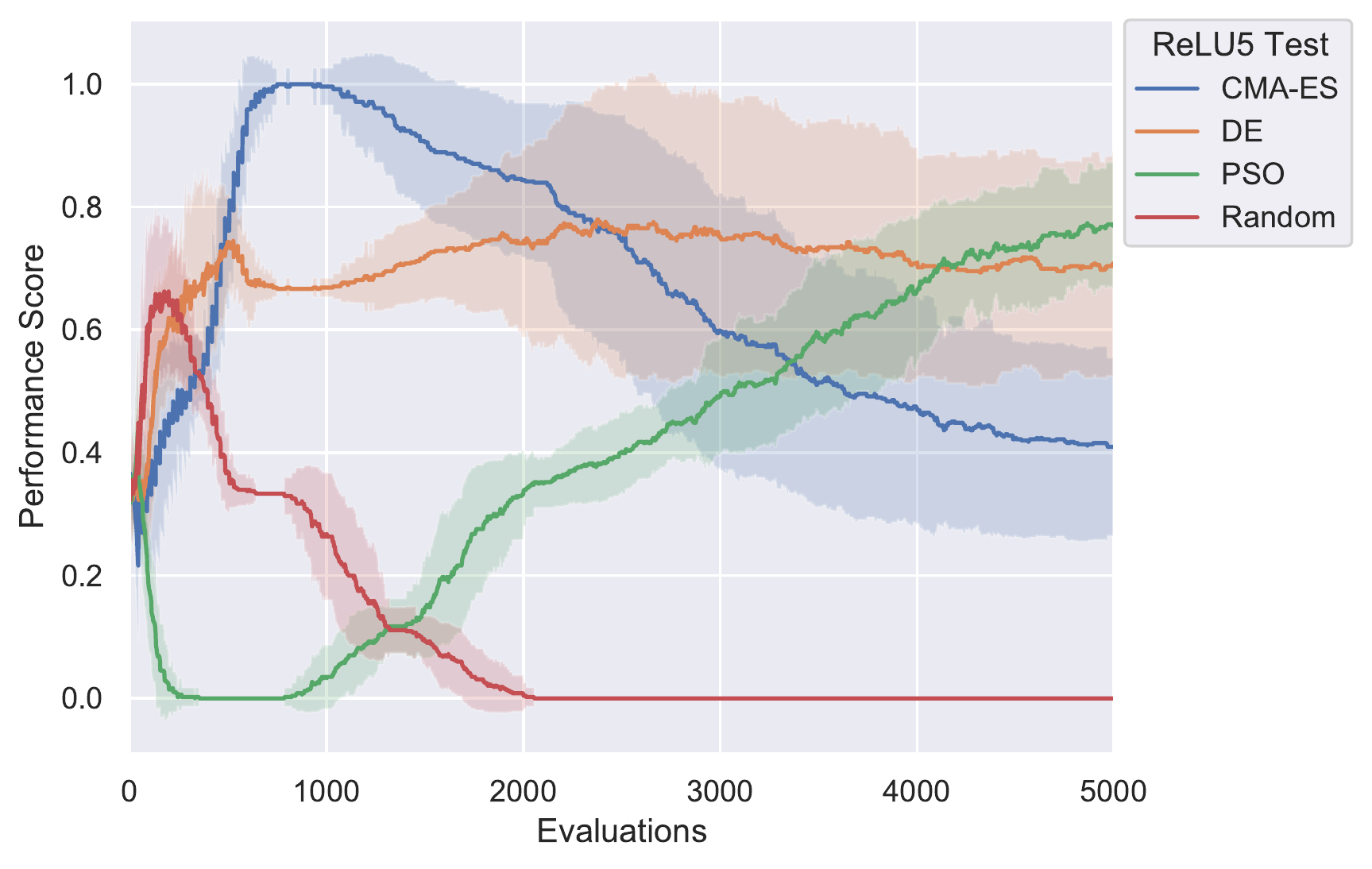}}
	\caption{Summarised relative performance on the test set of a 1-, 3- and 5-layer Tanh and ReLU models over $54$ function datasets. Lines denote mean performance with shaded bands denoting the standard deviation around the mean.}
	\label{tanh_relu_results}
\end{center}
\end{figure}

Figure \ref{tanh_relu_results} plots the normalised mean performance score over all $54$ problem instances for the full budget of $5000$ evaluations for the six NN models. Solid lines denote the mean performance score with shaded bands depicting the standard deviation around the mean.    
Two general observations from Figure \ref{tanh_relu_results} are that the three metaheuristics all performed significantly better than random search and that no single algorithm performed the best on all NN models. On the Tanh models (plots in the left column of Figure~\ref{tanh_relu_results}), CMA-ES performed the best on the 1-layer network, while DE performed the best on the 3- and 5-layer networks after the full budget of evaluations. On the ReLU models (plots in the right column of Figure \ref{tanh_relu_results}), PSO ultimately performed the best on all three models. CMA-ES was the quickest to find relatively good solutions, but converged to solutions that were inferior to those ultimately found by PSO.

\begin{figure}[!ht]
\begin{center}
\subfloat[1-layer Tanh model]{\includegraphics[width=0.49\textwidth]{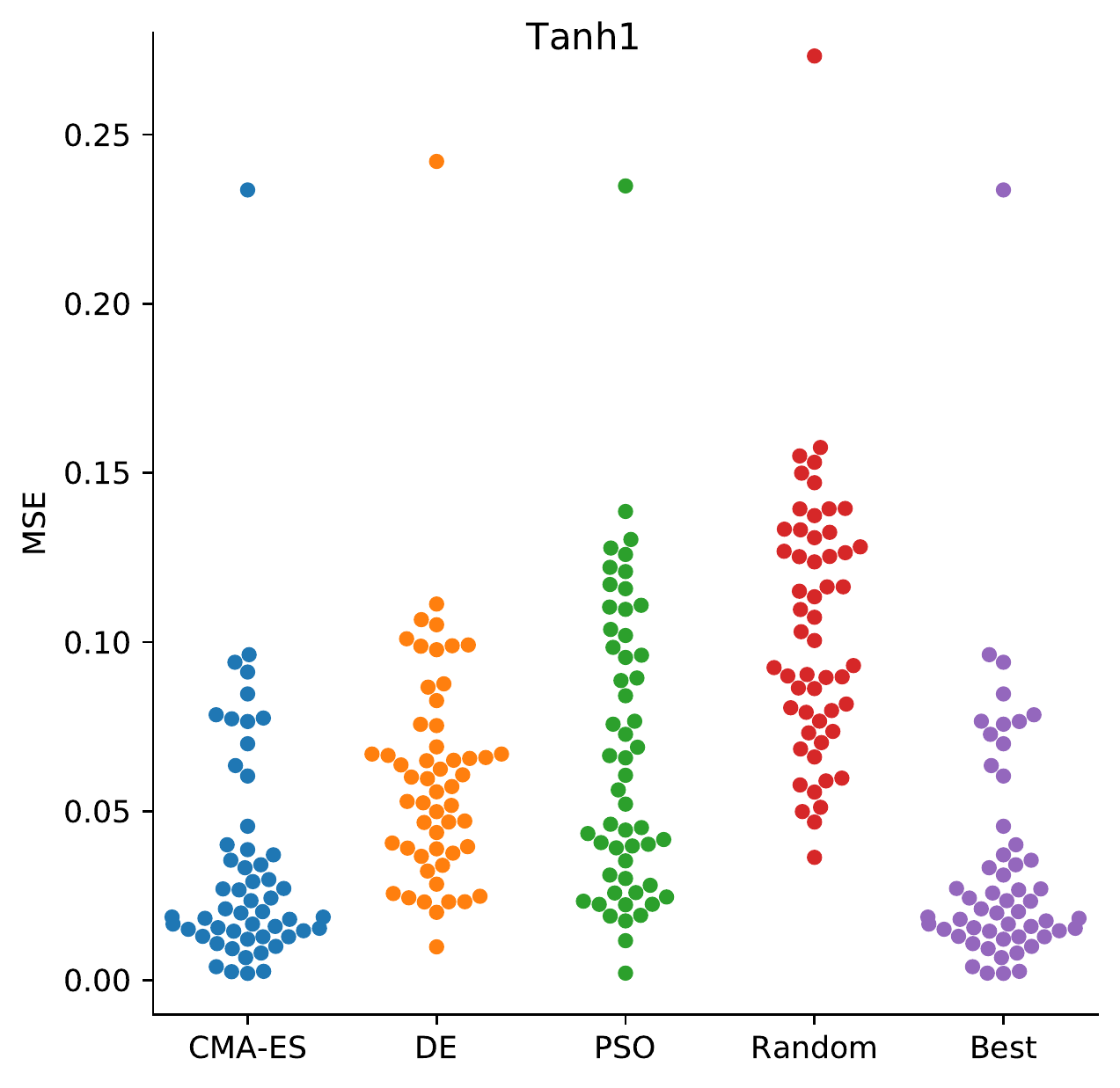}\label{fig:tanh1}} 
\subfloat[1-layer ReLU]{\includegraphics[width=0.49\textwidth]{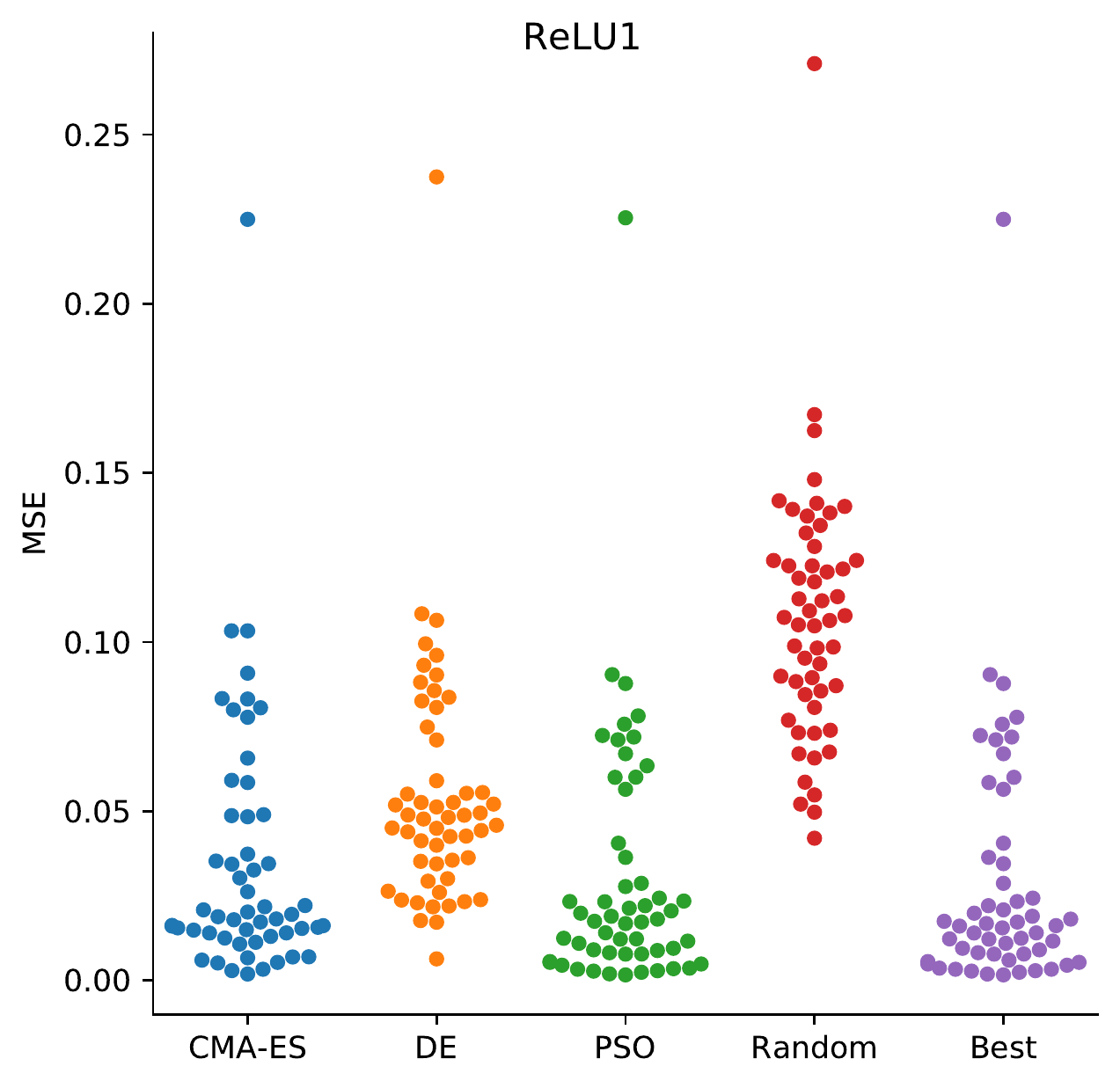}\label{fig:relu1}}\\
\subfloat[5-layer Tanh] {\includegraphics[width=0.49\textwidth]{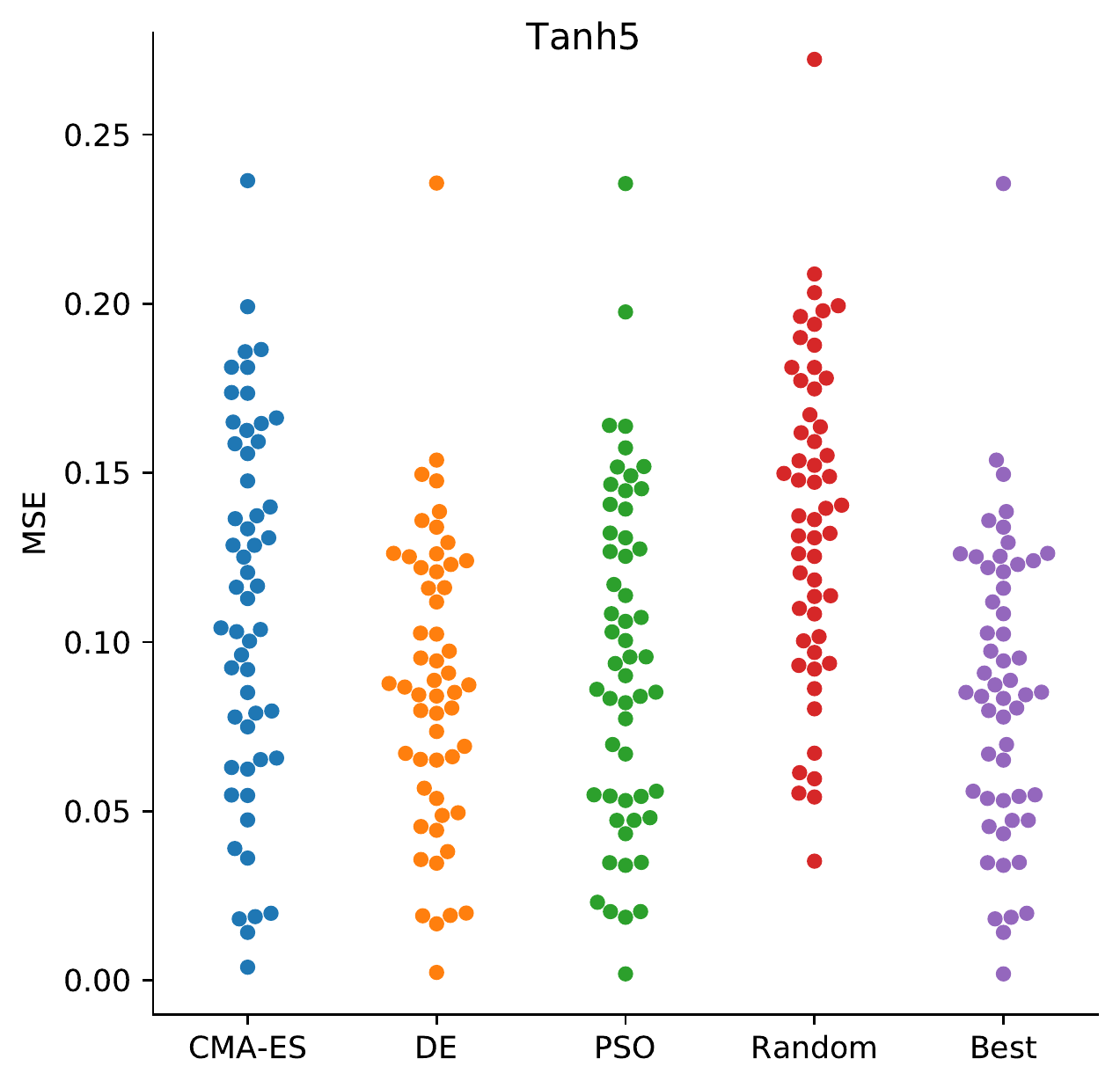} \label{fig:tanh5}} 
\subfloat[5-layer ReLU]
{\includegraphics[width=0.49\textwidth]{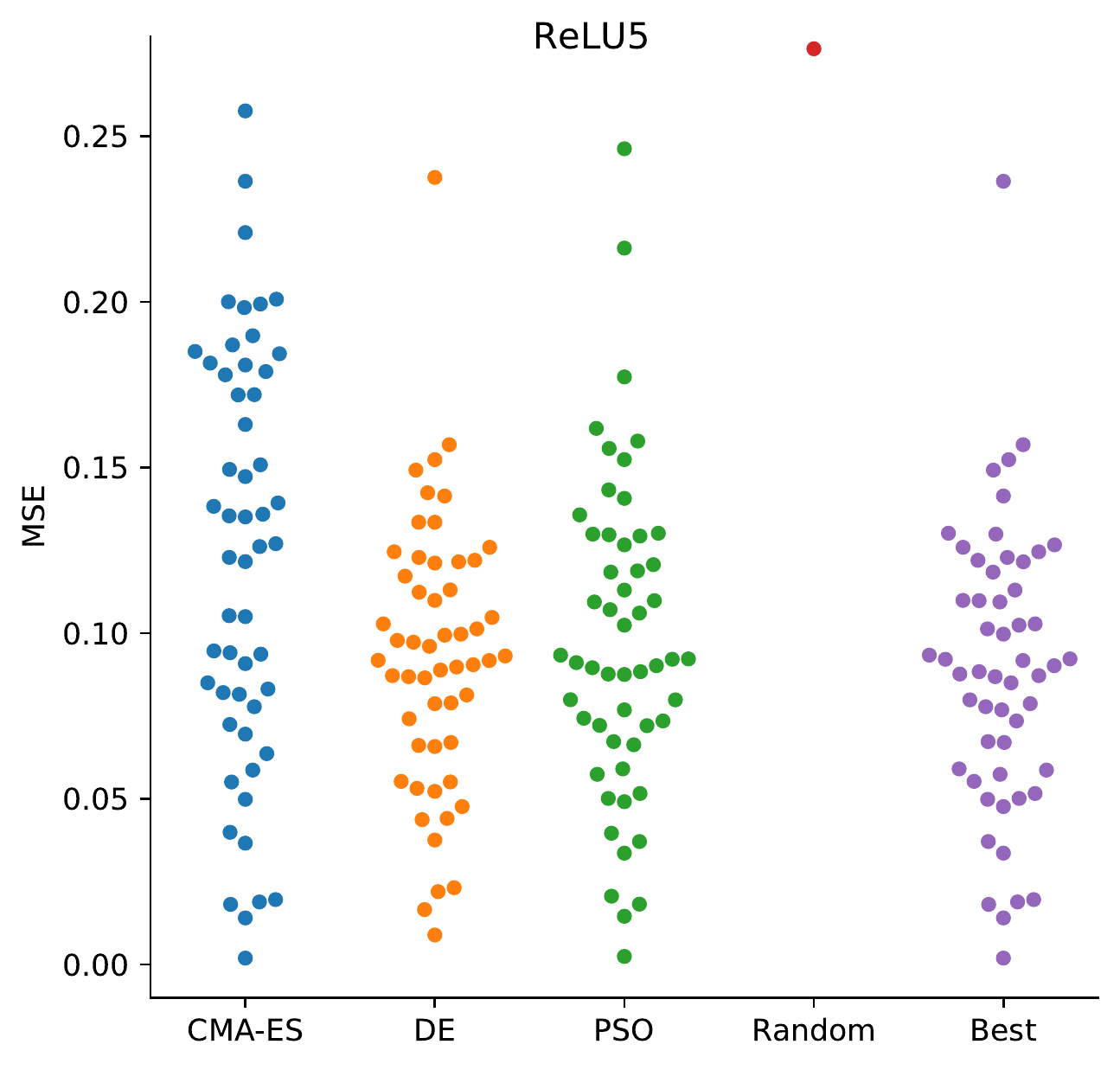}\label{fig:relu5}} 

\caption{Performance per function for each optimiser on two Tanh and ReLU models}
\label{fig:tanh_relu_both}
\end{center}	
\end{figure}

Figure \ref{fig:tanh_relu_both} visualises the performance of the algorithms on individual problem instances (the 3-layer models are omitted to save space). Each dot represents a single problem instance (54 of them in each column) and the vertical position corresponds to the average normalised MSE on the testing set from 30 runs of the algorithm after 5000 evaluations. Algorithms are plotted in a different colour for clarity.  The final column (purple dots) shows the results of the best performing algorithm for each problem instance. In these plots, better performing algorithms have a concentration of dots closer to the horizontal axis (lower MSE values). 

For all models, random search (red dots) performs worse than other algorithms. Contrasting Figure \ref{fig:tanh1} and \ref{fig:tanh5}, shows that for Tanh, the  performance across all algorithms deteriorates (fewer dots lower down) as the number of layers increase from 1 to 5, indicating an increase in problem difficulty. The same can be observed for ReLU architectures in Figures \ref{fig:relu1} to \ref{fig:relu5}. Note that in Figure \ref{fig:relu5}, only a single dot is shown for random search, because the other MSE values are above the range plotted on the graph. The very poor results of random search on ReLU5 indicates that these instances are more challenging overall than the other architectures.

\begin{figure}[!t]
\begin{center}
	\subfloat[Easom: easiest regression function]{%
		\includegraphics[width=0.48\textwidth]{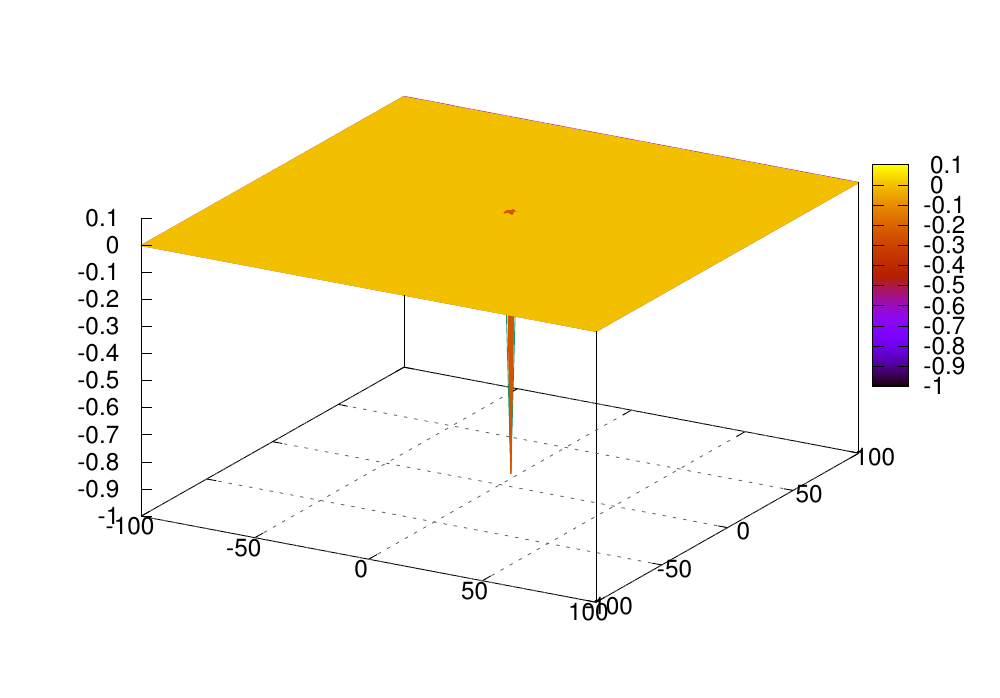}
	}
	\subfloat[Periodic: hardest regression function]{%
		\includegraphics[width=0.48\textwidth]{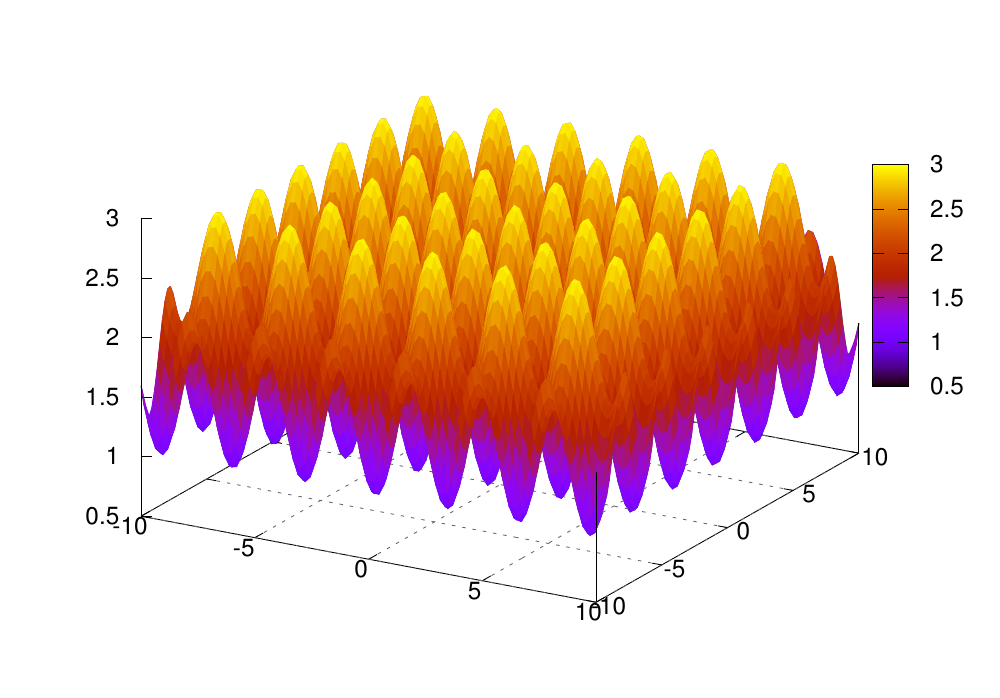}
	}
	\caption{Two regression tasks on which all algorithms performed the best and the worst.}
	\label{fig:easom_periodic}
	\end{center}
\end{figure}

Figure \ref{fig:tanh_relu_both} shows interesting outliers in the individual high dots at approximately 0.25 MSE. This corresponds to function 34, Periodic, on which most algorithms perform markedly worse than on the other problem instances. In contrast, the lowest dots in Figure \ref{fig:tanh5} correspond to function 20, Easom, on which all algorithms (except random search) achieved close to 0.00 MSE. Figure \ref{fig:easom_periodic} plots these two functions, clearly illustrating why it was easier for the search algorithms to fit models to Easom than to Periodic. 

\subsection{Analysis of Individual Problem Instances}
The next set of results highlights the range of difficulty of problems in the CORNN Suite for population-based algorithms in relation to a gradient-based approach. As a baseline, we provide the results of Adam, a form of gradient-descent with adaptive learning-rate that is popular for training deep NNs.

\begin{figure*}[!t]
	\centerline{\includegraphics[width=1\textwidth]{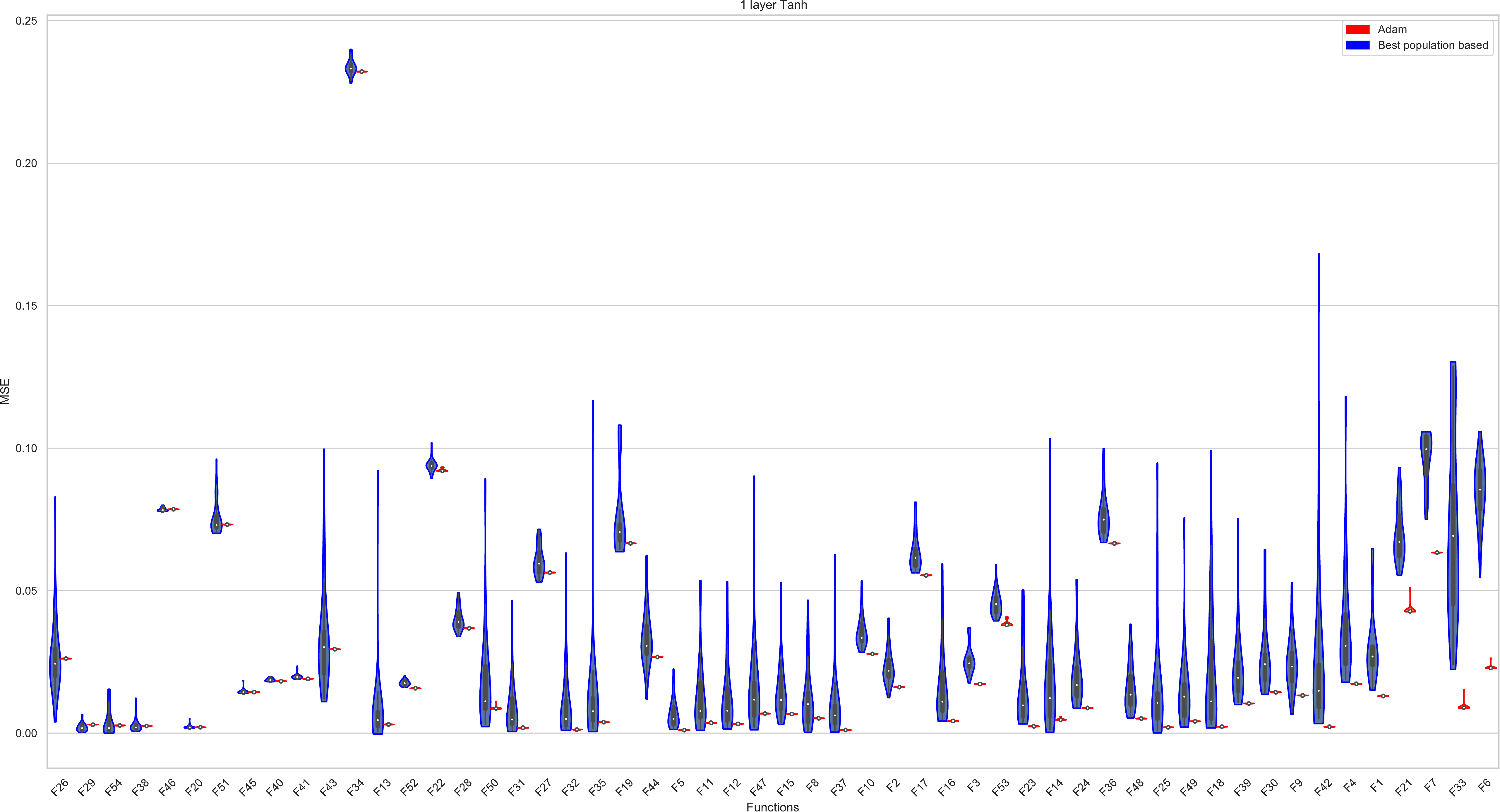}}
	\caption{Best performing population based optimiser compared to Adam on the 1-layer Tanh model}
	\label{fig:tanh1_popvsadam}
\end{figure*}

\begin{figure*}[!t]
	\centerline{\includegraphics[width=1\textwidth]{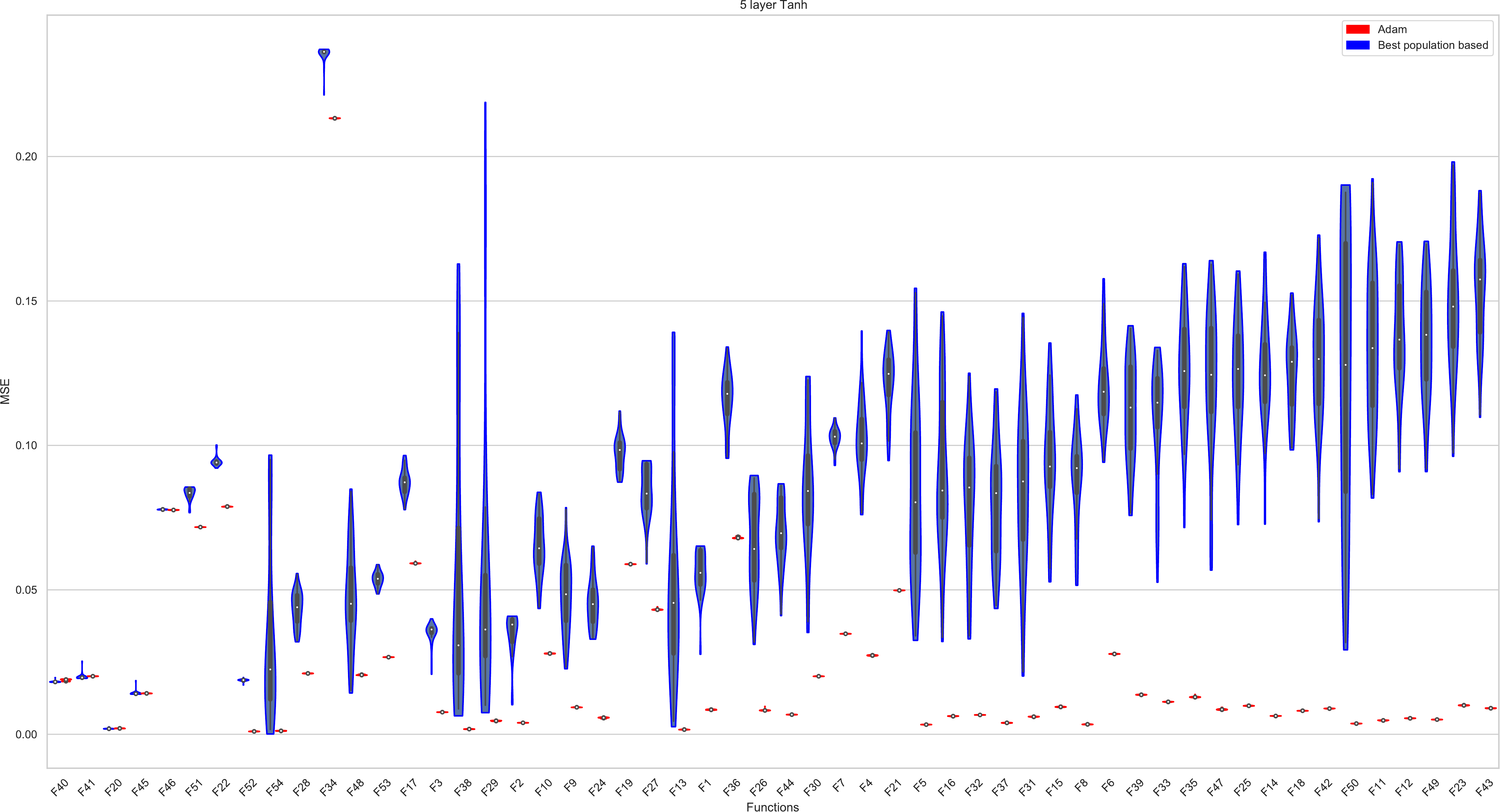}}
	\caption{Best performing population based optimiser compared to Adam on the 5-layer Tanh model}
	\label{fig:tanh5_popvsadam}
\end{figure*}

\begin{figure*}[!t]
	\centerline{\includegraphics[width=1\textwidth]{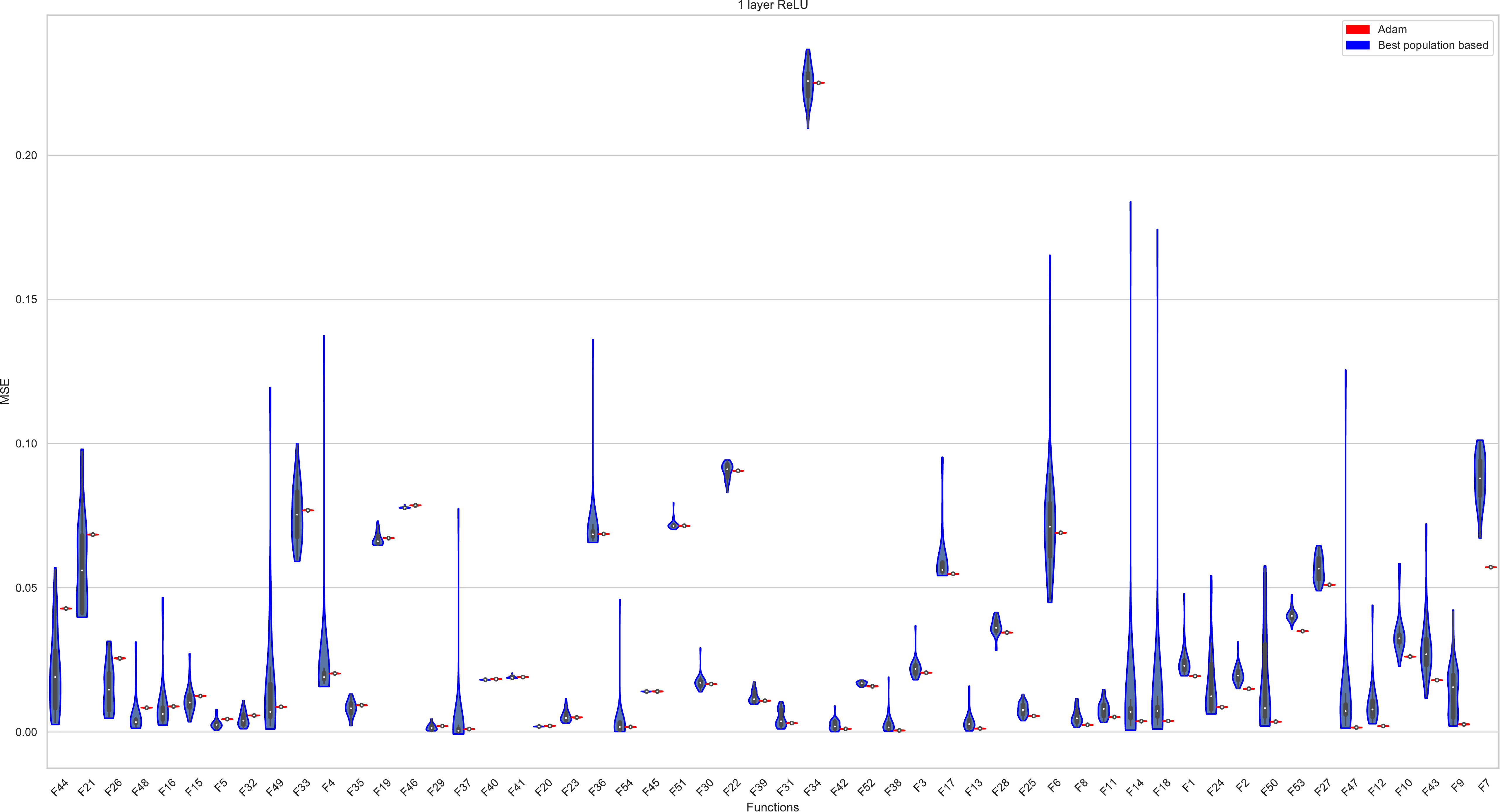}}
	\caption{Best performing population based optimiser compared to Adam on the 1-layer ReLU model}
	\label{fig:relu1_popvsadam}
\end{figure*}

\begin{figure*}[!t]
	\centerline{\includegraphics[width=1\textwidth]{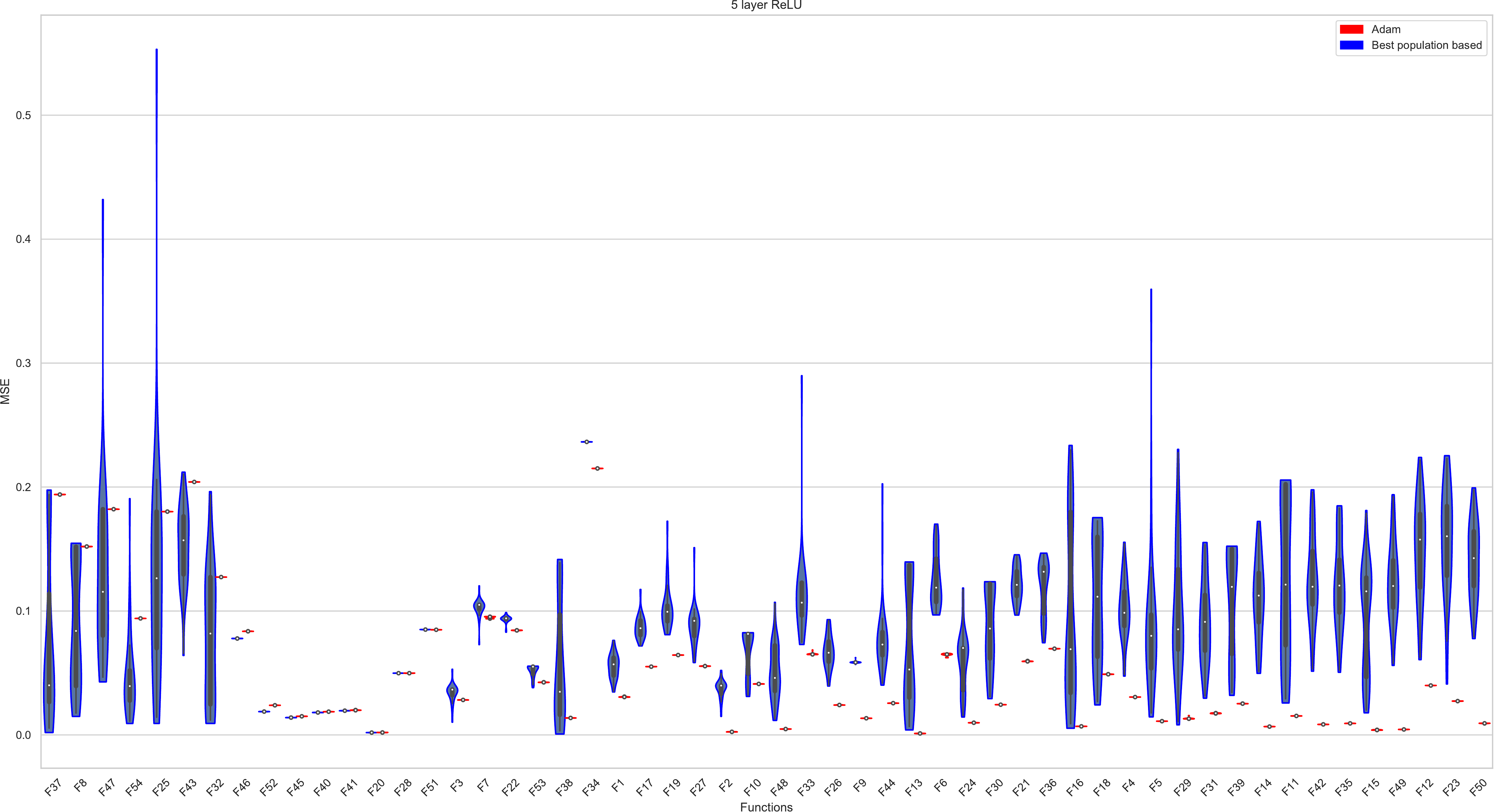}}
	\caption{Best performing population based optimiser compared to Adam on the 5-layer ReLU model}
	\label{fig:relu5_popvsadam}
\end{figure*}

Figures \ref{fig:tanh1_popvsadam} to \ref{fig:relu5_popvsadam} show violin plots of the distribution of MSE values from 30 runs on each of the 54 problem instances for Tanh1, Tanh5, Relu1 and Relu5, respectively. The blue violins represent the performance of the best population-based algorithm (of the three discussed in Section \ref{sec:algperf}) on the problem instance, while the red violins represent the performance of Adam. Note that most of the violins for Adam are very small due to the small variance in the performance over the 30 runs -- except for the random initial weights, Adam is a deterministic algorithm. The median MSE values appear as tiny white dots in the centre of each violin and the maximal extent of the violins are cropped to reflect the actual range of the data. 

The functions are sorted from left to right by the difference between the median MSE of the two approaches. For example, in Figure \ref{fig:tanh1_popvsadam}, the first function on the left is function 26 (Himmelblau), where the median MSE of the best population-based algorithm was slightly lower than the median MSE of Adam. From about the eighth function onwards, it can however be seen that Adam out-performed the best population-based algorithm. The superior performance of Adam is even more marked for the 5-layer Tanh model (Figure \ref{fig:tanh5_popvsadam}), with function 43 (Schwefel 2.22) resulting in the worst relative performance of the population-based algorithms. The most difficult function to fit for both Adam and population-based algorithms is evident by the high MSE values on function 34 (Periodic).


Figures \ref{fig:relu1_popvsadam} to Figure \ref{fig:relu5_popvsadam} show slightly better relative performance of the population based algorithms on the ReLU architectures compared to the Tanh architectures. On the left of Figure \ref{fig:relu5_popvsadam} we can see that the median MSE of the best population based algorithm is lower than Adam on about the first nine functions (37, 8, 47, 54, 25, 43, 46, and 52). 

In this way, these results provide a ranking of CORNN problem instances from the easiest to the hardest in terms of relative performance against gradient-based techniques. Future studies can focus on black-box algorithm development to reduce the gap in performance compared to gradient-based approaches. The CORNN Suite can be used in different types of analysis and not just in the way illustrated in this paper. For example, to simulate cases where the analytical gradient is not available for using gradient-based techniques, black-box optimisers can be benchmarked against one another to investigate the effectiveness on NN training tasks.

\section{Discussion}

The CORNN Suite complements existing benchmark sets with NN training tasks that can be used to benchmark the performance of any continuous black-box algorithm. Ideally, a benchmarking suite should be~\cite{BART2020}: (1)
diverse, (2) representative, (3) scalable and tunable, and (4) should have known solutions / best performance. The problem instances of the CORNN Suite are {\em diverse} as demonstrated by the wide range of performances by different algorithms. In addition, the problems are {\em representative} of real-world problems in  being continuous (which is more common in real-world settings than combinatorial problems~\cite{VAND2020}), with computationally expensive evaluation and involving the real-world task of NN training.
Problems are {\em scalable} and {\em tunable} through the selection of different NN models coupled with different regression tasks. 
For each problem instance the theoretical {\em optimal solution is known} (where MSE = 0 on the test set). However, given a fixed number of neurons in a model, we cannot rely on the universal approximator theorem \cite{HUAN2006} to guarantee the existence of an optimal solution of weights that will result in an error of zero. In addition to the theoretical minimum, we provide the performance of Adam as a baseline against which alternative algorithms can be benchmarked.   

\section{Conclusion}
The CORNN Suite is an easy-to-use set of unbounded continuous optimisation problems from NN training for benchmarking optimisation algorithms that can be used on its own or as an extension to existing benchmark problem sets. An advantage of the suite is that black-box optimisation algorithms can be benchmarked against gradient-based algorithms to better understand the strengths and weaknesses of different approaches.

The results in this paper provide an initial baseline for further studies. We have found that although Adam in general performed better, population-based algorithms did out-perform Adam on a limited set of problem instances. Further studies could analyse the characteristics of these instances using landscape analysis to better understand which NN training tasks are better suited to population-based approaches than gradient-based approaches. The CORNN Suite can also be used to try to improve population-based algorithms on NN training tasks. It would be interesting to analyse whether parameter configurations from tuning on the CORNN Suite can be transferred to other contexts to improve black-box metaheuristic algorithm performance on NN training tasks in general. 

\section*{Acknowledgments}
This work was supported by the National Research Foundation, South Africa, under Grant 120837. The authors acknowledge the use of the High Performance Computing Cluster of the University of South Africa. The authors also acknowledge the contribution of Tobias Bester for his initial implementation of the underlying regression functions.

%
%
%
\bibliographystyle{splncs04}
\bibliography{refs}
\end{document}